\documentclass[journal,hyphens]{IEEEtran}
\usepackage{amsmath,amsfonts}
\usepackage{amssymb}
\usepackage[backend=biber,style=ieee,maxnames=99,minnames=99]{biblatex}
\usepackage{pifont}
\usepackage{algorithmic}
\usepackage{algorithm}
\usepackage{array}
\usepackage{booktabs}
\usepackage{threeparttable}
\usepackage{multirow}
\usepackage[export]{adjustbox}
\usepackage{textcomp}
\usepackage{stfloats}
\usepackage{url}
\usepackage{listings}
\usepackage{color}
\usepackage{graphicx}
\usepackage[detect-none]{siunitx}
\usepackage[hidelinks]{hyperref}

\graphicspath{{images/}}

\DeclareSIUnit{\nothing}{\relax}
\DeclareSIUnit{\pixel}{px}
\DeclareSIUnit{\mac}{MAC}
\DeclareSIUnit{\frame}{frame}

\newcommand{\ie}{{i.e.},~}%
\newcommand{\eg}{{e.g.},~}%

\newcommand{\cmark}{\ding{51}}%
\newcommand{\xmark}{\ding{55}}%

\makeatletter
\newcommand{\todo}[1]{\noindent\textit{\color{red}\textbf{TODO}~#1}\@latex@warning{TODO: #1}} %LS
\makeatother

\newif\iffulldocument
\fulldocumenttrue
\makeatletter
\def\ps@IEEEtitlepagestyle{%
  \def\@oddhead{\hbox{}\hfil{\footnotesize This work has been accepted for publication in the IEEE RA-P journal. \copyright\ 2026 IEEE.}\hfil\hbox{}}%
  \def\@evenhead{\hbox{}\hfil{\footnotesize This work has been accepted for publication in the IEEE RA-P journal. \copyright\ 2026 IEEE.}\hfil\hbox{}}%
  \let\@oddfoot\@empty%
  \let\@evenfoot\@empty%
}
\makeatother

\begin{document}

\title{NanoCockpit: Performance-optimized Application Framework for AI-based Autonomous Nanorobotics}

\author{Elia~Cereda,~\IEEEmembership{Student Member,~IEEE,}
        Alessandro~Giusti,~\IEEEmembership{Member,~IEEE,}
        and~Daniele~Palossi~\IEEEmembership{Member,~IEEE}% <-this % stops a space
% \thanks{The authors would like to thank ....}% <-this % stops a space
\thanks{E. Cereda, A. Giusti, and D. Palossi are with the Dalle Molle Institute for Artificial Intelligence~(IDSIA), USI-SUPSI, 6962 Lugano, Switzerland. Corresponding author: {\tt\small elia.cereda@idsia.ch}.}% <-this % stops a space
\thanks{D. Palossi is also with the Integrated Systems Laboratory (IIS), ETH Z\"urich, 8092 Z\"urich, Switzerland.}%
%This work has been partially supported by \todo{XXXXXX under grant \#XXXXX}.}% <-this % stops a space
\thanks{The authors thank Rik Bouwmeester and Hanna M\"uller, for their support in making NanoCockpit part of the official Crazyflie software, and the colleagues at IDSIA that tested and used NanoCockpit in their research.}
\thanks{GitHub repository: \textit{\url{https://github.com/idsia-robotics/nanocockpit}}}% <-this % stops a space
}
% \markboth{Robotics and Automation Practice,~Vol.~\#, No.~\#, \#\#~2024}%
% {}

% \IEEEpubid{0000--0000/00\$00.00~\copyright~2024 IEEE}

\maketitle

\begin{abstract}
Autonomous nano-drones, powered by vision-based tiny machine learning (TinyML) models, are a novel technology gaining momentum thanks to their broad applicability and pushing scientific advancement on resource-limited embedded systems.
Their small form factor, i.e., a few tens of grams, severely limits their onboard computational resources to sub-\SI{100}{\milli\watt} microcontroller units (MCUs).
The Bitcraze Crazyflie nano-drone is the \textit{de facto} standard, offering a rich set of programmable MCUs for low-level control, multi-core processing, and radio transmission.
However, roboticists very often underutilize these onboard precious resources due to the absence of a simple yet efficient software layer capable of time-optimal pipelining of multi-buffer image acquisition, multi-core computation, intra-MCUs data exchange, and Wi-Fi streaming, leading to sub-optimal control performances.
Our \textit{NanoCockpit} framework aims to fill this gap, increasing the throughput and minimizing the system's latency, while simplifying the developer experience through coroutine-based multi-tasking.
In-field experiments on three real-world TinyML nanorobotics applications show our framework achieves ideal end-to-end latency, i.e. zero overhead due to serialized tasks, delivering quantifiable improvements in closed-loop control performance ($-$30\% mean position error, mission success rate increased from 40\% to 100\%).
\end{abstract}

\begin{IEEEkeywords}
% Official keywords: https://www.ieee-ras.org/publications/ra-p/keywords
Embedded Systems for Robotic and Automation,  Micro/Nano Robots, Hardware-Software Integration in Robotics 
% Hardware-Software Integration in Robotics, Micro/Nano Robots, Aerial Systems: Perception and Autonomy, AI-Enabled Robotics, Autonomous Vehicle Navigation
\end{IEEEkeywords}

\begin{refsection}
\section{Introduction} \label{sec:intro}

Autonomous nanorobotics is not only an attractive research field with many application scenarios, from human-robot interaction~\cite{pulp-frontnet,zhou23pedestrian} to narrow-space exploration~\cite{olejnik2020flappingservo,shao2022millipede,kabutz2023mclari,toumieh2024motion} and inspection~\cite{kumar2024watchers,zauli2024vibration}, but also one of the most challenging ``playgrounds'' to demonstrate new methodologies and technologies in the embedded systems domain~\cite{embeddedRobotics,goldberg2018legged,saeed2025flapping}.
On the one hand, the strict requirement of real-time execution of mission-critical tasks must match high-accuracy vision-based tiny machine learning (TinyML) perception workloads~\cite{pulp-frontnet,niculescu2021improving,lamberti2024imav,mengozzi2024drl,nanoflownet,navardi2023metae2rl,mueller2024bridges}.
On the other hand, these requirements must be fulfilled within a robot's form factor of a few 10s of grams with a stringent sub-\SI{100}{\milli\watt} onboard power budget, which leaves no other options than computationally limited microcontroller units (MCUs) with a few 100s \SI{}{\kilo\byte} memory and low-resolution vision sensors.

\begin{figure}[t]
  \centering
  \includegraphics[width=\linewidth]{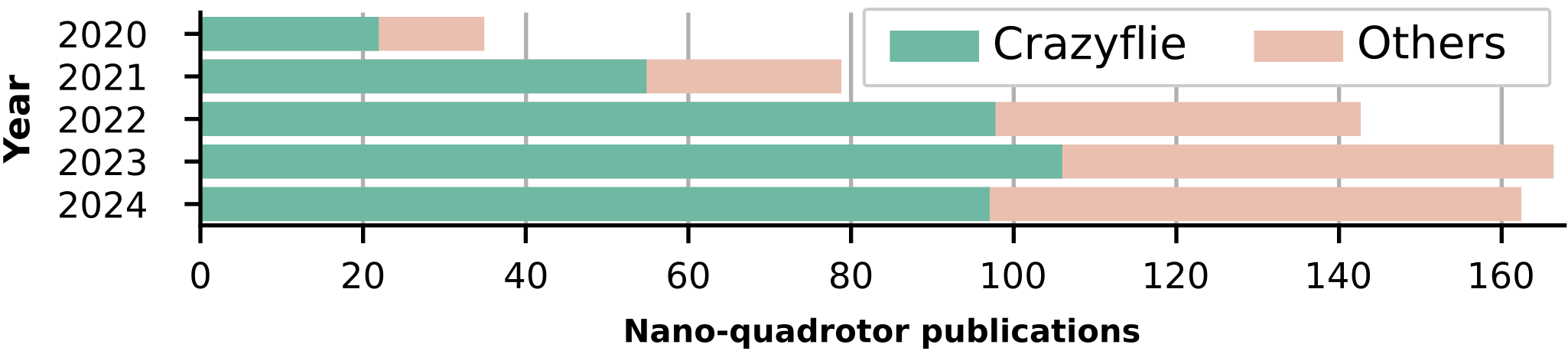}
  \caption{Scientific publications on nano-quadrotors over the last five years, according to the Elsevier Scopus citation database.}
  \label{fig:scopus}
\end{figure}

Among nanorobotics platforms, \emph{aerial} nano-robots are both the most technologically mature and the target of significant research interest.
% pose the strictest real-time requirements for closed-loop control.
We focus on the commercial Bitcraze Crazyflie, a \SI{27}{\gram} open-source nano-quadrotor built around a programmable STM32 MCU and augmented with the AI-deck expansion board~\cite{aideck}.
Figure~\ref{fig:scopus}, based on the Elsevier Scopus citation database\footnote{Elsevier Scopus only indexes scientific papers that underwent blind peer-review through a major publisher. \url{https://www.elsevier.com/products/scopus}}, witnesses how the Crazyflie is a \textit{de facto} standard employed in more than 65\% of nano-quadrotor publications from the last five years (additional details in Appendix~\ref{app:related-work}).
To date, the AI-deck is the State-of-the-Art (SotA) commercial solution for vision and computation aboard the Crazyflie and it provides an octa-core GWT GAP8 System-on-Chip (SoC), off-chip RAM/Flash memories, a low-resolution camera, and an Espressif ESP32 Wi-Fi module.
It enabled complex autonomous applications~\cite{sexton2022tbb,kazim2024nmpc,kalenberg2024stargate,lamberti2024imav} and served as blueprint for later academic open-hardware designs, e.g., the GAP9Shield~\cite{muller2024gap9shield}, which increase computation and memory.

\begin{table*}[t]
    \centering
    \caption{Performance assessment of SotA TinyML workloads running on the Crazyflie nano-drone with the GAP8 SoC. 
    }
    \label{tab:related_work}
    \resizebox{\textwidth}{!}{
        \begin{threeparttable}
        \begin{tabular}{
            r@{~}l
            c
            c
            S[table-format=2]
            S[table-format=4]
            S
            S[table-format=3]
            c
            S[table-format=2.1]
            S[table-format=2.1]
            c
        }
        \toprule
          \multicolumn{2}{c}{\multirow{2}[2]{*}{\textbf{Work}}}
        & {\multirow{2}[2]{*}{\textbf{Year}}}
        & {\multirow{2}[2]{*}{\textbf{TinyML task}}}
        & {\multirow{2}[2]{*}{\shortstack[c]{\textbf{Operations} \\ {[MMAC]}}}}
        & {\multirow{2}[2]{*}{\shortstack[c]{\textbf{Params} \\ {[\si{\kilo\nothing}]}}}}
        & {\multirow{2}[2]{*}{\shortstack[c]{\textbf{GAP8 efficiency} \\ {[MAC/cycle]}}}}
        & {\multirow{2}[2]{*}{\shortstack[c]{\textbf{GAP8 freq.} \\ {[\si{\mega\hertz}]}}}}
        % & {\multirow{2}[2]{*}{\shortstack[c]{\textbf{Running} \\ \textbf{onboard}}}}
        & {\multirow{2}[2]{*}{\shortstack[c]{\textbf{Compute}}}}
        & \multicolumn{2}{c}{\textbf{Throughput} [Hz]} 
        & {\multirow{2}[2]{*}{\shortstack[c]{\textbf{Performance} \\ {\textbf{drop} [\%]}}}} \\
        \cmidrule{10-11}
        & & & & & & & & & {Inference} & {Closed loop} & \\
        \midrule
            PULP-DronetV2 & \cite{niculescu2021improving}             & 2021 & lane follow + obst. avoid. & 41  & 320 & 4.5  & 175 & onboard & 19.0 & 12.8 & $\mathbf{-33}^{*}$ \\
            Navardi \textit{et al.} & \cite{navardi2022optimization}  & 2022 & lane follow + obst. avoid. & 40  & 98  & {--} & {--} & remote  & 25.0 & 9.5 & $\mathbf{-62}^*$ \\
            % Zhu \textit{et al.} & \cite{zhu22gesture}                 & 2022 & gesture recognition        & 36  & 364 & 4.0* & 150 & onboard & 16.7 & 9.1  & $\mathbf{-46}$ \\
            Zhou \textit{et al.} & \cite{zhou23pedestrian}            & 2023 & pedestrian tracking        & 36  & 220 & 4.8  & 250 & onboard & 33.3 & 28.0 & $\mathbf{-16}^\dagger$ \\
            Chen \textit{et al.} & \cite{chen23pedestrian}            & 2023 & pedestrian tracking        & 24  & 300 & 5.2  & 250 & onboard & 55.0 & 43.0 & $\mathbf{-22}^\dagger$ \\
            Pourjabar \textit{et al.} & \cite{pourjabar23multi}       & 2023 & obstacle avoidance         & 12  & 84  & 4.3  & 175 & onboard & 62.9 & 5.0  & $\mathbf{-92}^*$ \\
            NanoFlowNet & \cite{nanoflownet}                          & 2023 & optical flow               & 39  & 171 & 2.5  & 230 & onboard & 11.0 & 5.5  & $\mathbf{-50}^{\dagger\ddagger}$ \\
            \vspace{0.4em} % Add space before the next line (heading)
            Sartori \textit{et al.} & \cite{sartori2025autonomous}    & 2025 & object detect. + obst. avoid. & 1200  & 4350 & {--}  & {--} & remote & 19.6 & 8.4  & $\mathbf{-57}^*$ \\
            \multicolumn{4}{l}{\textbf{Deployed with our framework}} \\
        \midrule
            PULP-Frontnet & \cite{pulp-frontnet}                     & 2021 & human pose estimation       & 14 &  304 & 3.8  & 175 & onboard & 48.0 & 48.0 & $0$ \\
            Cereda \textit{et al.} & \cite{cereda2023secure}         & 2023 & human pose estimation       & 90 & 2228 & {--} & {--} & remote  & 40.0 & 40.0 & $0$ \\
            Crupi \textit{et al.} & \cite{crupi2024fcnn}             & 2024 & drone-to-drone localization &  8 &    8 & 1.8  & 175 & onboard & 39.0 & 39.0 & $0$ \\
            Lamberti \textit{et al.} & \cite{lamberti2024imav}       & 2024 & nano-drone racing           & 25 &  331 & 4.1  & 175 & onboard & 30.0 & 30.0 & $0$ \\
         \bottomrule
        \end{tabular}
        \begin{tablenotes}\footnotesize
        \item[*] Reported by the authors. $\dagger$ Derived from metrics reported by the authors. $\ddagger$ Our measurements on open-source code.
        \end{tablenotes}
        \end{threeparttable}
    }
\end{table*}

Roboticists working on fully autonomous nano-drones~\cite{niculescu2021improving,pulp-frontnet}, as well as those employing radio-connected remote computers to prototype sophisticated perception/control algorithms~\cite{navardi2022optimization,cereda2023secure}, need maximum computation and communication throughputs and minimal latency to achieve the best real-world application performance.
These require \textit{i)} time-optimal concurrent task execution, \eg camera acquisition and TinyML inference, \textit{ii)} efficient parallel multi-core processing, and \textit{iii)} zero-overhead data exchange, both inter-MCU and over the radio.
Until now, the Crazyflie platform lacked an integrated software layer to optimize task execution and data exchange across its MCUs, resulting in serialized execution, idle computational resources, and communication overheads.

To address this challenge, we present our novel \textit{NanoCockpit} framework, a set of performance-optimized plug-and-play libraries for AI-based autonomous nanorobotics, including \textit{i}) coroutine-based multi-tasking for asynchronous concurrent tasks; \textit{ii}) high-throughput camera drivers (GAP8), for multi-buffer acquisition up to \SI{150}{frame/\second} at $160\times\SI{160}{\pixel}$; \textit{iii}) a zero-copy Wi-Fi communication stack (ESP32) for real-time bi-directional communication at \SI{55}{\milli\second} mean round-trip latency.

To clarify the importance and need of our framework, Table~\ref{tab:related_work} surveys several SotA works~\cite{niculescu2021improving, navardi2022optimization, chen23pedestrian, zhou23pedestrian, pourjabar23multi, nanoflownet,sartori2025autonomous}.
We introduce the \textit{inference} and \textit{closed-loop} throughput metrics; the former considers only the TinyML workload, while the latter includes all tasks (e.g., camera acquisition, inference, communication).
The two throughputs match in an ideal implementation,  while all surveyed works fall short of delivering the full potential closed-loop throughput.
Some works identify system limitations, e.g., sub-optimal pipelining of image acquisition and processing~\cite{niculescu2021improving,pourjabar23multi} and Wi-Fi streaming overhead~\cite{navardi2022optimization,sartori2025autonomous}, as the cause of 33\%--92\% drops~\cite{niculescu2021improving,pourjabar23multi} in closed-loop throughput, compared to the sole inference throughput.
The remaining works~\cite{chen23pedestrian}, \cite{zhou23pedestrian}, \cite{nanoflownet} only report their closed-loop throughput.
Still, in Table~\ref{tab:related_work}, we derive the inference throughput from intermediate metrics reported by the authors, i.e., number of multiply-and-accumulate (MAC) operations per inference, computational efficiency (MAC/cycle), and GAP8 clock frequency, showing performance drops between 16\% and 50\%.
Through NanoFlowNet~\cite{nanoflownet}'s open-source code, we further reproduce its measurements independently and attribute the closed-loop drop to serialized camera acquisition, lack of double buffering, and sub-optimal camera configuration, all aspects addressed by our framework.

The second part of Table~\ref{tab:related_work} highlights the impact of our framework.
Several TinyML models~\cite{pulp-frontnet, cereda2023secure, crupi2024fcnn, lamberti2024imav}, when deployed on top of NanoCockpit, reach the optimal closed-loop throughput, bounded only by unavoidable inference time.
For these models, we provide three in-field experiments in Section~\ref{sec:results}, to analyze the real-world impact of our multi-MCU framework.
On all experiments, we improve the final closed-loop performance, as much as $-30\%$ in mean position error and a mission success increase from 40\% to 100\%.
We believe open-sourcing our work and our technical insights delivers a significant resource to the nanorobotics community.
\section{Hardware-software architecture} \label{sec:method}

\subsection{Robotic platform}

\begin{figure}[t]
  \centering
  \includegraphics[width=\columnwidth]{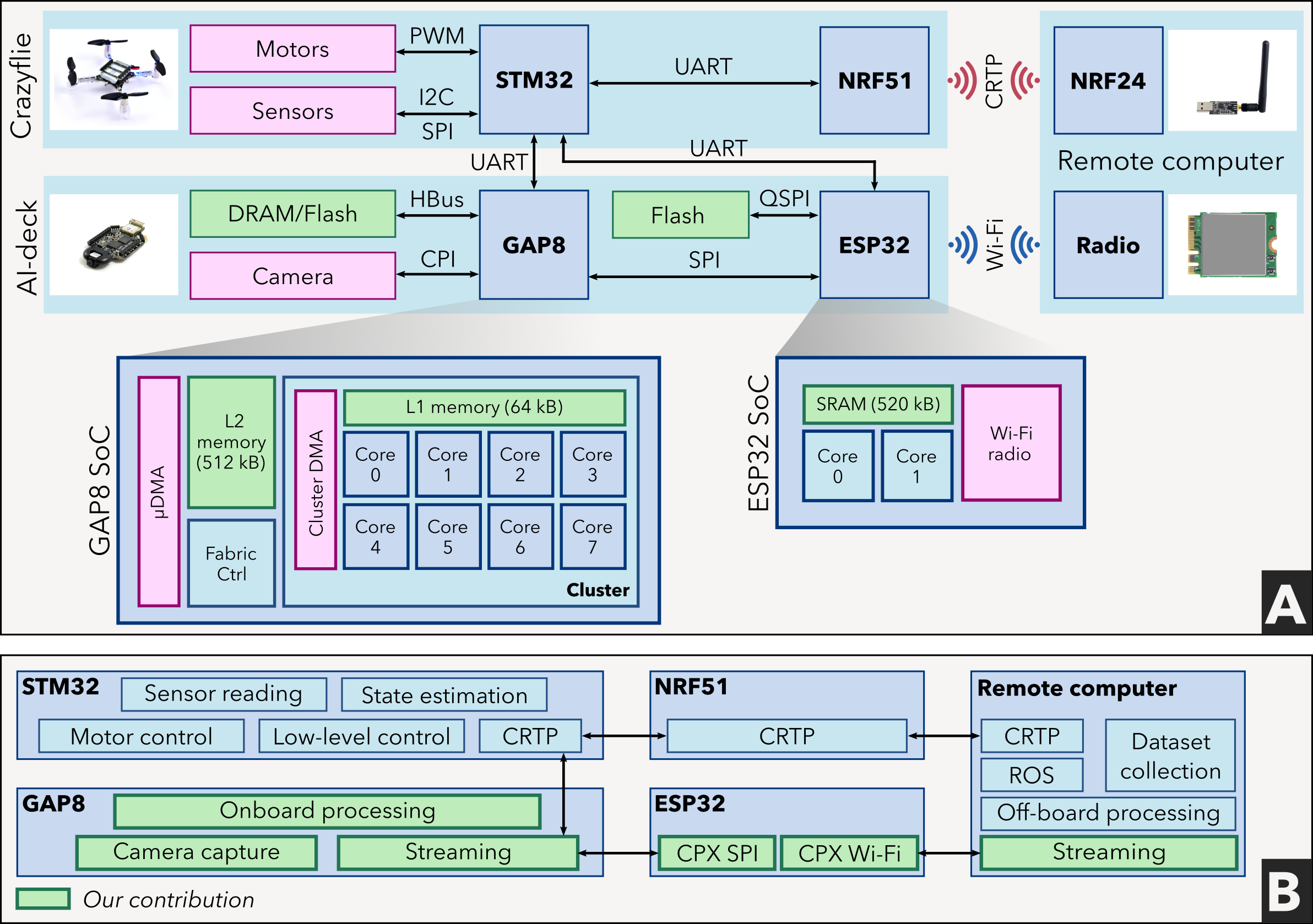}
  \caption{Hardware (A) and software (B) overviews of the Crazyflie nano-drone with the AI-deck companion board and a remote computer.}
  \label{fig:cps}
\end{figure}

\textbf{Hardware.}
Figure~\ref{fig:cps}-A shows the hardware design of the Crazyflie extended with the AI-deck board and a radio-connected remote computer.
On the nano-drone, an STM32 MCU manages flight-control tasks, while a Nordic NRF51 handles the \SI{2.4}{\giga\hertz} radio.
The AI-deck features a QVGA Himax camera (HM01B0), a GWT GAP8 octa-core RISC-V-based SoC, off-chip HyperDRAM/Flash of 8 and \SI{64}{\mega\byte}, respectively, and an Espressif ESP32 SoC for Wi-Fi connectivity.
Figure~\ref{fig:cps}-A also depicts the internal architecture of the GAP8 and ESP32.
GAP8 is composed of two general-purpose power domains, a single-core \textit{fabric controller} (FC) for data-handling tasks and an octa-core \textit{cluster} (CL) for computational-intensive workloads, such as vision-based TinyML algorithms~\cite{pulp-frontnet,bompani23bio}.
GAP8 also features two on-chip memories, i.e. a \SI{512}{\kilo\byte} L2 memory and a \SI{64}{\kilo\byte} L1 low-latency scratchpad, a micro direct memory access ($\mu$DMA) unit to interact with off-chip peripherals, e.g., camera and memories, and a cluster DMA for data movement between on-chip memories.
The ESP32 features a 32-bit dual-core Xtensa LX6 CPU, a Wi-Fi radio, \SI{520}{\kilo\byte} on-chip SRAM and \SI{2}{\mega\byte} off-chip Flash memory.

Remote communication is possible over two wireless channels: a low-latency and low-bandwidth Crazy Real-Time Protocol (CRTP) driven by the NRF51 and a high-bandwidth Crazyflie Packet eXchange (CPX)~\cite{bitcraze2025cpx} protocol, through the ESP32.
The former typically transfers control setpoints and logging, while the latter is meant for high-volume data exchange, such as image streaming.
The STM32 communicates over two universal asynchronous receiver-transmitter (UART) interfaces, with the NRF51 and with the GAP8. 
Instead, the GAP8 exchanges data with the ESP32 over serial peripheral interface (SPI), with the camera over camera parallel interface (CPI), and with off-chip memories over HyperBus.

\textbf{Software.}
Figure~\ref{fig:cps}-B illustrates the software tasks that typically run on the various MCUs aboard a Crazyflie and on a remotely connected computer, highlighting which tasks have been contributed by our framework.
The typical execution flow employs the STM32 for all the drone's basic functionalities, such as stabilization and control, and state estimation.
The GAP8 SoC typically handles computationally intensive tasks: image acquisition, processing, e.g., vision-based TinyML workloads, and optionally streaming to the remote computer of images and/or onboard processing results.
The ESP32 performs CPX routing between the SPI and Wi-Fi interfaces to feed the remote computer that can collect data and perform off-board computation, optionally leveraging the Robot Operating System (ROS)~\cite{macenski2022ros2}.
As low-level software stacks, the STM32 and ESP32 use the FreeRTOS real-time operating system (RTOS)~\cite{barry2024freertos} with full pre-emptive multitasking, while the GAP8 runs a lightweight runtime, called PMSIS.
Our framework extends the PMSIS functionalities, offers optimized camera drivers, and improves the Wi-Fi stack.

\begin{figure}[t]
  \centering
  \includegraphics[width=\linewidth]{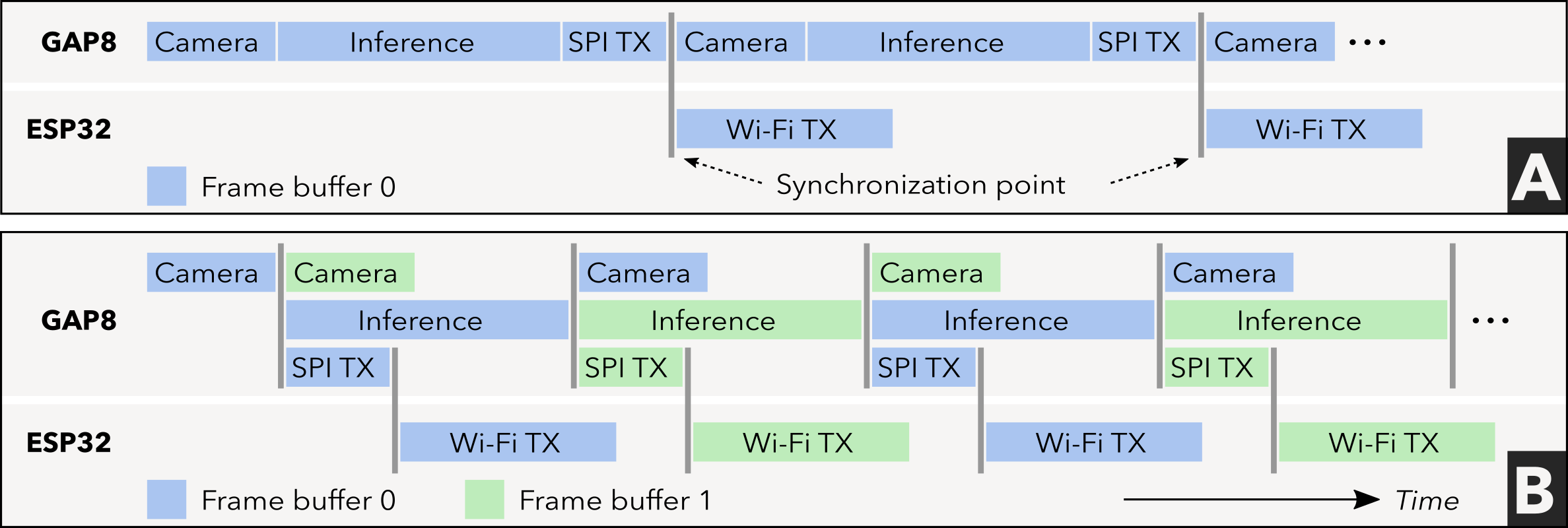}
  \caption{(A) Serialized execution without our NanoCockpit framework vs. (B) pipelined (double-buffered) execution with our framework.}
  \label{fig:pipeline}
\end{figure}

\subsection{Framework architecture} \label{sec:framework}

\textbf{Coroutine-based multi-tasking.}
The ideal closed-loop throughputs from Table~\ref{tab:related_work} are achieved when all computational units are always loaded with useful work, i.e., zero idle time.
In our context, camera acquisition and communication must overlap with GAP8's computation, e.g., TinyML inference, by employing DMA-based data transfers and multi-tasking execution.
However, GAP8's PMSIS runtime is a single-thread hardware abstraction layer and only supports asynchronous execution via event-based callbacks.
This programming model requires significant effort to implement cooperative behavior and forces programmers to explicitly handle all events and their synchronization~\cite{belson2019coroutines}.
The control flow of each task ends up spread among many callback functions.
So far, this limitation prevented fine-grained task overlapping and forced several SotA works~\cite{niculescu2021improving,nanoflownet} to resort to suboptimal serialized execution, where camera capture, inference, and SPI transmission (TX) tasks execute one after the other, as Figure~\ref{fig:pipeline}-A.

Our NanoCockpit framework extends PMSIS with a custom asynchronous and cooperative programming layer based on stackless co-routines~\cite{belson2019coroutines}, allowing the programmer to easily suspend a task and resume it when a desired condition is met.
In Figure~\ref{fig:pipeline}-B, we show an example of the fully pipelined execution enabled by our framework.
Each concurrent task is implemented as a distinct co-routine that can use synchronous control flow constructs (e.g., \texttt{if}, loops), suspend itself, and yield control to another task (details in Appendix~\ref{app:coroutine}).
The proposed mechanism is extremely fast, with sub-\SI{10}{\micro\second} context switches (on the order of a function call) and, compared to a full RTOS such as FreeRTOS on STM32, it has 8$\times$ lower memory overhead (i.e., \SI{18}{\byte} per task vs. $\ge$\SI{150}{\byte} required by each task's stack in FreeRTOS).

\textbf{Multi-buffered camera drivers.}
Our Himax camera has two operative modes: a \textit{trigger} mode, where the GAP8 acts as master, requesting single images (up to $\sim$\SI{30}{\frame/\second}), and a \textit{streaming} mode, where the camera is programmed once and then produces a continuous stream of images.
The streaming mode allows for the highest frame rate (up to \SI{150}{\frame/\second} with our framework), but the GAP8 must work synchronously with the camera and be ready to store a new image every time it is produced.
However, the camera's stock drivers and GAP8 runtime do not support this optimized execution flow. 
Our framework provides optimized camera drivers (fine-grained timing control), allowing the camera to operate at the ideal frame rate for the application and a software layer for pipelined multi-buffered acquisition, preventing frame dropping and jittering.
In Figure~\ref{fig:pipeline}-B, we show an example of double-buffered image acquisition, where the $\mu$DMA acquires new images in parallel with the CL's inference and SPI TX.

\textbf{Zero-copy Wi-Fi stack.}
The original Wi-Fi stack based on the CPX protocol suffers from high latency, serial-only execution, and lack of congestion control~\cite{bitcraze2025cpx}.
We provide a novel communication stack that fulfills the CPX specifications while enabling high-throughput and low-latency bi-directional Wi-Fi streaming between the drone and a remote computer.
Our stack, which spans the STM32, GAP8, and ESP32 MCUs, also offers hardware-based timestamping for precise synchronization of data acquired across different onboard MCUs, which is fundamental for dataset collection.
Compared to the original CPX stack, our version implements zero-copy packet transmission on GAP8 and a multi-tasking and multi-buffer router implementation on ESP32, enabling complete overlapping of SPI and Wi-Fi transfers, as shown in Figure~\ref{fig:pipeline}-B.
Our stack can stream $160\times\SI{160}{\pixel}$ images at $\SI{72}{\hertz}$, i.e., $2.4\times$ higher than the original CPX.
\section{Results} \label{sec:results}

In this section, we explore three different real-world applications, and to assess their quantitative figures, we employ a motion capture system (additional details in Appendix~\ref{app:experiments}).

\subsection{Human pose estimation} \label{sec:d2h_results}

A convolutional neural network (CNN) takes as input a $160\times\SI{96}{\pixel}$ image and outputs the subject's pose relative to the drone's horizontal frame.
The poses are filtered by a Kalman filter and used by a closed-loop velocity controller to keep the drone at a fixed distance in front of the subject, i.e., \SI{1.5}{\meter}.
To enrich our analysis, we evaluate two CNN models: PULP-Frontnet~\cite{cereda2021improving} running onboard and the MobileNetV2-based CNN from Cereda et al.~\cite{cereda2023secure}, running on a remote computer.
PULP-Frontnet has \SI{304}{\kilo\nothing} parameters and requires \SI{14.3}{\mega\mac} per inference, while the MobileNetV2 has $7\times$ more parameters and requires \SI{90}{\mega\mac} operations per inference.

\begin{table}[t]
    \centering
    \caption{Human pose estimation control performance.}
    \label{tab:d2h_results}
    \resizebox{\columnwidth}{!}{
        \footnotesize
        \renewcommand{\arraystretch}{1.03} % Just to make space for "Onboard" in the first column
        \begin{tabular}{cccccc}
        \toprule
        & \multirow{2}[2]{*}{\textbf{Model}} &
        \multirow{2}[2]{*}{\shortstack[c]{\textbf{Throughput} \\ {[\si{\hertz}]}}} &
        \multirow{2}[2]{*}{\shortstack[c]{\textbf{End-to-end} \\ \textbf{latency} {[\si{\milli\second}]}}} & 
        \multicolumn{2}{c}{\textbf{Control error}} \\
        \cmidrule(lr){5-6}
            & & & & $e_{xy}$ [\si{\meter}] & $e_{\theta}$ [\si{rad}] \\
        \midrule
        \parbox[t]{2mm}{\multirow{3}{*}{\rotatebox[origin=c]{90}{{Onboard}}}} &
        \multirow[c]{3}{*}{\shortstack[c]{PULP-Frontnet \\ \cite{cereda2021improving}}}
             & 12 & 30.3 & 0.96 & 0.63 \\
           & & 24 & 30.3 & 1.01 & 0.69 \\
           & & 48 & 30.3 & \textbf{0.80} & \textbf{0.50} \\
        \midrule
        \parbox[t]{2mm}{\multirow{6}[2]{*}{\rotatebox[origin=c]{90}{{Remote}}}} &
        \multirow[c]{3}{*}{\shortstack[c]{Cereda \textit{et al.}~\cite{cereda2023secure}}}
                                        & 10 & 136.6       & 0.83 & 0.55 \\
                                     &  & 20 & 146.2       & 0.74 & 0.46 \\
                                     &  & 40 & 168.6       & \textbf{0.65} & \textbf{0.41} \\
        \cmidrule(lr){2-6}
                                     & 
        \multirow[c]{3}{*}{\shortstack[c]{Cereda \textit{et al.}~\cite{cereda2023secure} \\ w/ \SI{500}{\milli\second} delay}}
                                        & 10 & 636.6 & 0.87 & 0.55 \\
                                     &  & 20 & 646.2 & 0.91 & 0.50 \\
                                     &  & 40 & 668.6 & \textbf{0.81} & \textbf{0.46} \\
        \bottomrule
        \end{tabular}
    }
\end{table}

\begin{figure}[t]
  \centering
  \includegraphics[width=\linewidth]{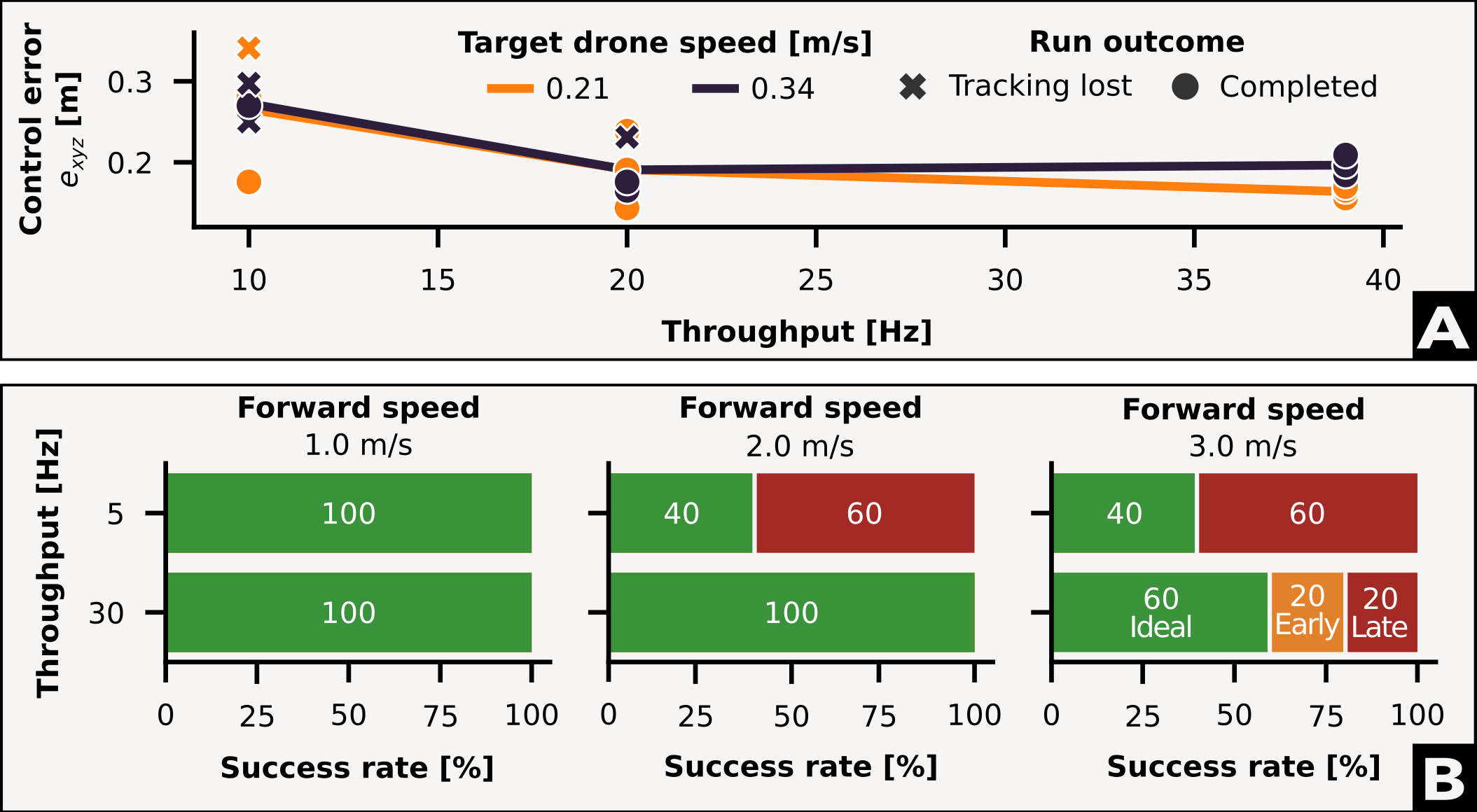}
  \caption{Closed-loop control performance in the A) drone-to-drone localization and B) nano-drone racing obstacle avoidance experiments.}
  \label{fig:d2d_imav_results}
\end{figure}

A human subject, not part of training data, walks on a predefined path while the nano-drone predicts its pose and follows it.
For each CNN, we test three different throughput configurations, and for each, we run three tests.
The throughput represents how often a new camera acquisition and inference is triggered, while the end-to-end latency measures the time from the image acquisition to the CNN's output reaching the low-level controller.
To assess the impact of increased throughput on the final control errors, ideally, we would like to keep constant end-to-end latency for all configurations.
However, due to Wi-Fi congestion, the MobileNetV2 is affected by unpredictable end-to-end latencies.
To decouple throughput from end-to-end latency, we introduce a third experiment with the MobileNetV2 model, where we inject an additional latency of \SI{500}{\milli\second} while keeping the throughput unchanged.

Table~\ref{tab:d2h_results} reports the mean horizontal position ($e_{xy}$) and mean angular ($e_{\theta}$) errors.
For each model, the highest throughput consistently scores the lowest errors, while, among all models, the peak performance comes with the biggest CNN, the higher throughput, and the lower latency.
In fact, the off-board MobileNetV2 running at \SI{40}{\hertz} achieves a 19\% lower error despite a more than \SI{100}{\milli\second} latency compared to the PULP-Frontnet and scores almost on-par when affected by a high latency, \ie $>$\SI{600}{\milli\second}.
This trend suggests that high-throughput inference, with more predictions per second for the filtering process, reduces the impact of noise on predictions and leads to better final performance.

\subsection{Drone-to-drone localization}

Our Crazyflie (called \textit{observer}) is tasked to estimate the relative pose of a target peer nano-drone in its field of view.
A fully convolutional neural network~\cite{crupi2024fcnn} (FCNN) takes in input a $160\times\SI{160}{\pixel}$ image from the onboard camera and produces three $20\times\SI{20}{\pixel}$ probability maps, that are processed to extract the target's image-plane coordinates $(u, v)$, and distance $d$.
The target drone performs a predefined 10-meter spiral trajectory in front of the observer drone, which uses the FCNN and the controller onboard to track and follow it.
We reproduce the trajectory over either 48 or \SI{24}{\second} (average target drone speeds of 0.21 and \SI{0.34}{\meter/\second}, respectively).
At each speed, we test three end-to-end throughputs (10, 20, and \SI{39}{\hertz}), i.e., accounting from the image acquisition to the final pose forwarded to the low-level controller.
Each test is repeated three times.

Figure~\ref{fig:d2d_imav_results}-A shows that higher throughputs reduce both the mean position error $e_\mathit{xyz}$, i.e. Euclidean distance in 3D space, and the instances of target tracking loss (\pmb{$\times$} in the plot).
The benefits of an increased throughput are clear for both target speed configurations: at \SI{0.21}{\meter/\second}, we lose track of the target only once at \SI{10}{\hertz}, while at \SI{0.34}{\meter/\second}, we lose track of the target in 2 out of 3 runs at \SI{10}{\hertz} and only one time at \SI{20}{\hertz}.
Only at the highest throughput, the observer drone never loses track of the target, achieving the best $e_\mathit{xyz}$ error of \SI{0.18}{\meter}.

\subsection{Nano-drone racing}

Our obstacle avoidance task is based on a CNN for a drone racing scenario~\cite{lamberti2024imav}, which is fed with onboard $162\times\SI{162}{\pixel}$ images and predicts three probabilities of collision $[0,1]$: left, center, and right vertical splits of the image.
The model has \SI{331}{\kilo\nothing} parameters and requires \SI{25}{\mega\mac} operations per frame, achieving up to \SI{30}{\hertz} throughput on the GAP8 with our framework.
The collision probabilities are filtered by a low-pass filter and used to drive the drone's forward speed.

The drone takes off at \SI{4}{\meter} from the obstacle and flies toward it until the collision probability reaches a threshold, breaking until stopping.
In the first \SI{2}{\meter}, the predictions are always at 0, allowing the controller to reach the desired max speed.
If the drone stops between 0.15 and \SI{2}{\meter} from the obstacle, we consider the test successful (ideal) or failed otherwise (too early/late).
In Figure~\ref{fig:d2d_imav_results}-B, we explore three max speeds (1, 2, and \SI{3}{\meter/\second}) and two CNN throughputs either at \SI{5}{\hertz} or \SI{30}{\hertz}, coupled with 0.05 and 0.30 breaking threshold, respectively.
For each configuration, we repeat five runs.
As the velocity grows, the advantage of higher throughputs becomes evident, with a 60\% success rate at the highest speed.
\section{Conclusion} \label{sec:conclusion}

This work presents the NanoCockpit framework for increasing the throughput and minimizing the system's latency while simplifying the programmability of the Crazyflie nano-drone.
Real-world experiments demonstrate how increased throughput and reduced perception-to-control latency are fundamental sources of improved behavior of the nano-drone in all use cases.
Finally, we open-source our framework for the benefit of the wider research community.

%\section*{Acknowledgements}

\begin{table*}[p]
    \centering
    \caption{Description of supplementary materials}
    \label{tab:supplementary_material}
    \resizebox{\textwidth}{!}{
        \begin{threeparttable}
        \begin{tabular}{
            c
            m{4in}
        }
        \toprule
            \textbf{Supplementary material} & 
            \textbf{Description} \\
        \midrule
            \multirow{9}{*}{\shortstack[c]{GitHub repository\\\url{https://github.com/idsia-robotics/nanocockpit}}} & 
            Source code of our NanoCockpit framework: \\
            &
            \begin{itemize}
                \item Framework code for the GAP8, ESP32, and STM32 MCUs and the remote computer (ROS and Python).
                \item Standalone examples of our co-routine and synchronization primitives, CPX communication stack, camera acquisition and streaming (Section~\ref{sec:framework}).
                \item Full application deployed on top of our framework: PULP-Frontnet CNN~\cite{cereda2021improving} and closed-loop controller for the human pose estimation experiment (Section~\ref{sec:d2h_results}).
                \item Profiling and debugging tools: GAP8 and ESP32 event trace collection (over GPIO or UART), Wireshark protocol dissectors to inspect CPX Wi-Fi traffic.
                \item Manually filtered and annotated dataset from the literature review in Figure~\ref{fig:scopus}.
            \end{itemize} \\
            \multirow{4}{*}{\texttt{supplementary-materials.pdf}} & 
            Additional methodology and implementation details: \\
            & 
            \begin{itemize}
                \item Appendix\,\ref{app:related-work}: Methodology, Scopus queries, and analysis of our literature review.
                \item Appendix\,\ref{app:coroutine}: Implementation details of our co-routine primitive.
                \item Appendix\,\ref{app:experiments}: Detailed descriptions of our experiment tasks, neural network deployment, experimental setup, and the evaluation metrics.
                \item Appendix\,\ref{app:discussion}: Discussion on NanoCockpit relevance for other heterogeneous cyber-physical systems.
            \end{itemize} \\
         \bottomrule
        \end{tabular}
        \end{threeparttable}
    }
\end{table*}
\clearpage

\printbibliography
\end{refsection}

% \vfill

\iffulldocument
    \clearpage
    
    \begin{refsection}
    \appendices
    \section{Nanorobotics survey} \label{app:related-work}

\begin{table*}[t]
    \centering
    \caption{Survey on nano-quadrotors appearing in peer-reviewed scientific publications in the last five years.}
    \label{tab:nano-drone-survey}
    \resizebox{\linewidth}{!}{
        \renewcommand{\arraystretch}{1.05}
        \begin{tabular}{lccccccccrrrrrr}
        \toprule
        \multirow{2}[2]{*}{\shortstack[c]{\textbf{Nano-drone}}} &
        \multirow{2}[2]{*}{\shortstack[c]{\textbf{Dimensions} \\ {$L \times W \times H$ [\si{\milli\meter}]}}} &
        \multirow{2}[2]{*}{\shortstack[c]{\textbf{Weight} \\ {[\si{\gram}]}}} &
        \multirow{2}[2]{*}{\shortstack[c]{\textbf{Battery} \\ {[\si{\milli\ampere\hour}]}}} &
        \multirow{2}[2]{*}{\shortstack[c]{\textbf{Flight time} \\ {[\si{\minute}]}}} &
        \multirow{2}[2]{*}{\shortstack[c]{\textbf{SoC}}} &
        % \multirow{2}[2]{*}{\shortstack[c]{\textbf{Sensors}}} &
        % \multirow{2}[2]{*}{\shortstack[c]{\textbf{Motors}}} &
        \multirow{2}[2]{*}{\shortstack[c]{\textbf{Expansion} \\ \textbf{interface}}} &
        \multirow{2}[2]{*}{\shortstack[c]{\textbf{Open} \\ \textbf{firmware}}} &
        \multirow{2}[2]{*}{\shortstack[c]{\textbf{Open} \\ \textbf{hardware}}} &
        \multicolumn{6}{c}{\textbf{Publications}} \\
        \cmidrule{10-15}
         & & & & & & & & & 2020 & 2021 & 2022 & 2023 & 2024 & \textbf{Total} \\
        \midrule
        \textbf{Bitcraze Crazyflie}    & $92\times92\times29$ & 27 &  250 &  7 & STM32F405 & \cmark & \cmark & \cmark & 22 & 55 & 99 & 107 & 98 & 381 \\ % https://www.bitcraze.io/products/crazyflie-2-1-plus/, brushed
        \textbf{DJI Tello}             & $98\times93\times41$ & 80 & 1100 & 13 & Intel Movidius Myriad 2 & \xmark & \xmark & \xmark & 2 & 17 & 29 & 38 & 42 & 128 \\ % https://www.ryzerobotics.com/tello/specs, https://tellopilots.com/threads/the-difference-between-regular-tello-and-tello-edu-with-photos.2734/, brushed, https://fccid.io/2AOOEWM0041801
        \textbf{Parrot Mambo}          & $160 \times 78 \times 9.8$ & 62 & 660 & 10 & Parrot P6 (ARM9) & \xmark & \xmark & \xmark & 4 & 2 & 12 & 17 & 15 & 50 \\ % https://www.crutchfield.com/S-JkawTMLTM1A/p_333F727006/Parrot-Mambo-FPV-Drone.html, MCUs https://www.youtube.com/watch?v=oP80ehn4HdA, https://www.mdpi.com/aerospace/aerospace-10-00512/article_deploy/html/images/aerospace-10-00512-g013.png, https://fccid.io/2AG6IDELOS3/Internal-Photos/Internal-Photos-Internal-View-Parrot-Mambo-3078447, Parrot P6 https://github.com/parrot-opensource/mambo-opensource/blob/master/sources/linux-2.6.36/atom.mk
        \textbf{Parrot Rolling Spider} & $140 \times 140 \times 36$ & 57 & 550 & 7 & Parrot P6 (ARM9) & \xmark & \xmark & \xmark & 2 & 1 & 1 & 1 & 0 & 5 \\ % https://www.droneflyers.com/parrot-rolling-spider-minidrone-first-look-review-rating/, MCU ARM9 https://www.pinterest.com/pin/live-teardown-of-the-parrot-rolling-spider-mini-drone-by-ifixit-on-the-arm-booth-at-maker-fair-silicon-valley-2015--197032552425136291/, https://fccid.io/RKXDELOS1/Internal-Photos/Internal-Photos-2585281, Parrot P6 https://github.com/parrot-opensource/rollingspider-opensource/blob/master/sources/linux-unknown/atom.mk, dimensions https://www.bhphotovideo.com/c/product/1079565-REG/parrot_pf723002_parrot_rolling_spider_drone_red.html
        \textbf{ESPcopter}             & $90\times90\times35$ & 35 & 240 & 7 & ESP8266 & \cmark & \cmark & \cmark & 0 & 0 & 2 & 1 & 1 & 4 \\ % https://espcopter.com, brushed, height from product images
        \textbf{Syma x27 Ladybug}      & $103\times103\times26$ & -- & 200 & 7 & -- & \xmark & \xmark & \xmark & 0 & 1 & 0 & 1 & 0 & 2 \\ % https://www.symatoys.com/goodshow/x27-syma-x27-whole-of-the-world-at-your-fingertip.html, brushed, https://fccid.io/QV7GC8875253
        \textbf{Arcade PICO}           & $90\times90\times30$ & 30 & 150 & 5.5 & -- & \xmark & \xmark & \xmark & 1 & 1 & 0 & 0 & 1 & 3 \\ % https://www.pama.com/wholesale/ARCPCM20, dimensions https://www.amazon.co.uk/Arcade-Pico-Drone-Altitude-Hold/dp/B074NCT2KV, weight estimated from dimensions+battery 
        \textbf{Cheerson CX-10W}       & $32\times32\times22$ & 17 & 150 & 4 & STM32F050K & \xmark & \xmark & \xmark & 1 & 0 & 0 & 0 & 0 & 1 \\ % https://www.firstquadcopter.com/reviews/cheerson-cx-10w-review/, MCU https://www.edn.com/teardown-a-tiny-camera-drone/, https://avdweb.nl/tech-tips/electronics/quadcopter
        % \textbf{Flapping Wing, but na& & & & & & & & & & no} & 0 & 0 & 0 & 1 & 0 \\
        \textbf{Pluto X}                & $160\times160\times45$ & 60 & 600 & 9 & STM32F303 & \cmark & \cmark & \cmark & 1 & 0 & 0 & 0 & 0 & 1 \\ % https://www.dronaaviation.com/plutox/, height estimated from product image
        \textbf{RadioShack DIY Drone}   & -- & -- & 250 & 5 & -- & \xmark & \xmark &  \xmark & 1 & 0 & 0 & 0 & 0 & 1 \\ % https://www.hobbytown.com/radioshack-diy-drone-starter-kit-rsh2770422/p788054
        \textbf{Custom}                 & -- & -- & -- & -- & -- & \xmark & \xmark & \xmark & 0 & 1 & 1 & 1 & 4 & 7 \\
        \bottomrule
        \end{tabular}
    }
\end{table*}

Table~\ref{tab:nano-drone-survey} surveys the most popular nano-quadrotors used in research, and provides the raw information used to build Figure~\ref{fig:scopus}, in Section~\ref{sec:intro}.
The most widely used nano-drone is the Bitcraze Crazyflie\footnote{https://www.bitcraze.io/products/crazyflie-2-1-plus/}, referenced in 381 publications over the last five years (65\%), for which we provide a detailed description of its architecture in Section~\ref{sec:method}-A. 
The second most used is the DJI Tello\footnote{https://store.dji.com/en/product/tello} (and TelloEDU) nano-quadrotor, which accounts for 21\% of publications.
The Parrot Mambo\footnote{https://www.parrot.com/en/support/documentation/mambo-range} and its variant, the Parrot Rolling Spider, account together for 9\% of publications in the same five years.
Compared to the Crazyflie, these are all slightly larger drones and come with closed-source firmware and hardware.
Nonetheless, they allow for the development of off-board algorithms that can be tested via their Python SDKs.

All remaining nano-drones rarely appear in research, with just 1 to 5 publications each, over five years.
Among these, the ESPcopter and Pluto X nano-quadrotors are the most interesting as they are based on the ESP8266 and the STM32F303 microcontroller unit (MCU), respectively.
Both nano-drones share Crazyflie's modular design and offer fully open-source firmware, making them attractive alternatives for education and research.
However, to date, they do not have onboard cameras or additional computational resources, limiting their applicability either to simple algorithms or requiring off-board processing.
The Syma X27 Ladybug, Arcade PICO, Cheerson CX-10W, and the RadioShack DIY Drone Kit are all consumer-oriented toy drones with limited adoption in research, despite their attractive ultra-small form factor, e.g., the Cheerson is a $3\times\SI{3}{\centi\meter}$ ($\mathrm{length} \times \mathrm{width}$) tiny nano-drone.
Finally, a handful of publications propose their own custom-built solutions, showing limited research interest in developing novel nano-quadrotor platforms and preferring to leverage existing ones.

\begin{figure}[t]
    \centering
    \begin{lstlisting}[language=Python, caption={Elsevier Scopus query used for our nanorobotics survey.}, label={lst:scopus}, numbers=none]
TITLE-ABS-KEY(
  (
    (nano-uav OR nano-uavs OR
        nano-drone OR nano-drones OR 
        nanocopter OR nano-quadrotor) OR 
    (palm-size AND uav) OR 
    (palm-size AND drone) OR 
    (pocket-size AND drone) OR 
    (mini-drone) OR 
    (mini-uav) OR 
    (pocket-sized AND uav) OR 
    (palm-sized AND uav) OR 
    (pocket-sized AND drone) OR 
    (palm-sized AND drone) OR 
    (mini-uav)
  ) OR (  
    (Crazyflie) OR 
    (Tello AND (drone OR uav OR quadrotor)) OR 
    (ladybug OR (syma AND (ladybug OR x27)) AND 
        (drone OR uav OR quadrotor)) OR 
    (Mambo AND (drone OR uav OR quadrotor)) OR 
    ("Rolling Spider" AND 
        (drone OR uav OR quadrotor)) OR 
    (CrazePony2 AND (drone OR uav OR quadrotor)) OR 
    (ArduBee AND (drone OR uav OR quadrotor)) OR 
    (ESPcopter) OR 
    ("CX-10W" AND (drone OR uav OR quadrotor)) OR 
    (PlutoX AND (drone OR uav OR quadrotor)) OR 
    ("Arcade Pico" AND 
        (drone OR uav OR quadrotor)) OR 
    (RadioShack AND (drone OR uav OR quadrotor))
  )
) AND 
PUBYEAR > 2019 AND PUBYEAR < 2025 AND 
NOT DOCTYPE(cr) AND NOT DOCTYPE(er)
    \end{lstlisting}
\end{figure}

With these aerial nanorobotics platforms in mind, we query the Elsevier Scopus citation database (update date: 9 Jan 2026) to collect reliable metrics about their employment in scientific publications.
We retrieve the Scopus data for all nano-drone publications over the last five years, starting with the query in Listing~\ref{lst:scopus} and focusing on the years 2020--2024 to account for the $\sim$six month delay of the Scopus indexing process.
This first query results in 710 entries, which we complement with 325 other publications acknowledged by Bitcraze in their website\footnote{https://www.bitcraze.io/portals/research/}.
After removing duplicates and manually checking the fit w.r.t. our scope, i.e., surveying all and only publications employing nano-quadrotors, we obtain  the final 586 entries reported in Table~\ref{tab:nano-drone-survey}.
The final CSV file with our survey is provided on the NanoCockpit repository (available online).
    \clearpage
    \section{Co-routine implementation}
\label{app:coroutine}

\begin{figure*}[t]
  \centering
  \includegraphics[width=\linewidth]{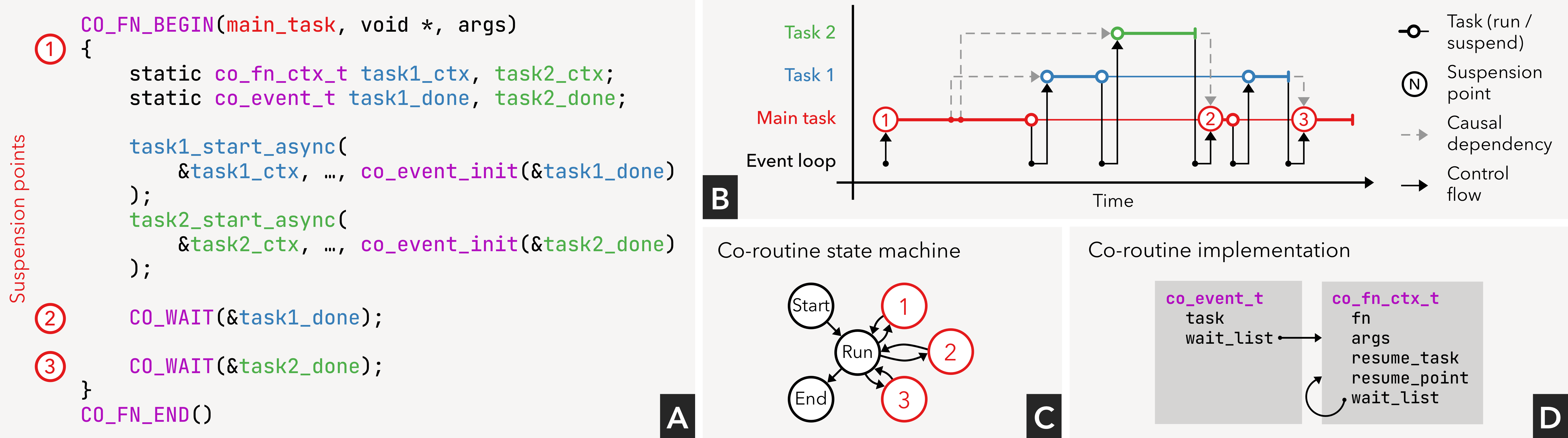}
  \caption{Co-routine-based cooperative multi-tasking. A) C API example of our co-routine implementation. B) Execution flow with interactions among tasks and event loop. C) Co-routine state machine with start, running, and end states and the three suspension points. D) Internal implementation of co-routine structures.}
  \label{fig:coroutines}
\end{figure*}

Stackless co-routines~\cite{belson2019coroutines} are structured asynchronous programming primitives that enable cooperative concurrency with extremely low memory and time overheads.
They represent a common programming pattern with a clear control flow that eases the development, improves readability, and fosters the reusability of code.
We provide a co-routine example in Figure~\ref{fig:coroutines}: a \texttt{main\_task} starts two sub-tasks, \ie \texttt{task1} and \texttt{task2}, in a fork-join scenario, which reflects common situations in an embedded application.
The two tasks execute concurrently in background while the \texttt{main\_task} suspends and waits for their completion.
This scenario matches, for example, the camera acquisition loop, which concurrently starts the \textit{inference} and streaming \textit{SPI TX} tasks (Figure~\ref{fig:pipeline}-A) for each frame and waits until both tasks complete before proceeding to the next frame.

Figure~\ref{fig:coroutines}-A demonstrates the syntax of our co-routine API.
A co-routine is delimited by the \texttt{CO\_FN\_BEGIN} and \texttt{\_END} macros, which define the co-routine's name (\eg \texttt{main\_task}), a pointer-sized parameter (\texttt{args}, 4 bytes), and its body.
Each \textit{co-routine instance} (\ie an invocation of a co-routine running at a given time) stores its state in an \textit{execution context} object, \texttt{co\_fn\_ctx\_t}.
As the main task starts two co-routine instances, \texttt{task1} and \texttt{task2}, in Figure~\ref{fig:coroutines}-A, it needs to reserve two execution contexts to store their respective states.
\textit{Event} objects, \texttt{co\_event\_t}, represent external events that a co-routine can wait upon.
An event is initialized with \texttt{co\_event\_init} and waited upon using the \texttt{CO\_WAIT} macro.
\texttt{CO\_WAIT} defines a \textit{suspension point} in the co-routine: when reached, it registers the co-routine in a linked list of \textit{waiters} stored by the event, then suspends the co-routine, storing the co-routine's state in its execution context.

The execution flow of our example is detailed in the temporal diagram of Figure~\ref{fig:coroutines}-B.
An event loop, provided by the PMSIS hardware abstraction layer, keeps track of events that have completed.
When the \texttt{task1} and \texttt{task2} co-routine are started, they're pushed on the event loop's ready list.
When the \texttt{main\_task} reaches a \texttt{CO\_WAIT} wait call, it yields control to the event loop, which then processes the next ready events in order.
This implements a cooperative multi-tasking system in which tasks must regularly yield to the event loop to ensure all events are processed on time.

The \texttt{task\#\_start\_async} functions are examples of \textit{asynchronous functions}, functions which start a background task that continues beyond the their own execution.
In PMSIS API conventions, these functions are identified by the \texttt{\_async} suffix appended to their name and take an \textit{event} object as their last parameter, to notify the caller of the background task's completion.
Our \texttt{co\_event\_t} event object can be passed to all PMSIS asynchronous functions, to allow our co-routines to wait on any background task from PMSIS drivers.

Figure~\ref{fig:coroutines}-D shows the internal details of the two key C structures, \texttt{co\_event\_t} and \texttt{co\_fn\_ctx\_t}.
Upon suspension, a co-routine stores the suspension point from which the co-routine will resume (\texttt{resume\_point}, \SI{2}{bytes}) when the waited-upon event is completed (\texttt{resume\_task}, \SI{4}{bytes} pointer).
The \texttt{resume\_point} identifies the current state of the co-routine, in the state machine from Figure~\ref{fig:coroutines}-C.
When a \texttt{co\_event\_t} is completed, its list of waiters is walked through by the event loop and all waiting co-routines are resumed, restarting their execution from their respective \texttt{resume\_point}.

Notably, our implementation does not keep a private stack for each co-routine, saving memory, which is a critical resource in any MCU-class device.
A single stack, owned by the main task, is shared across all co-routines, with each co-routine accessing it while actively running.
However, the lack of a private stack limits the use of local variables across suspension points.
A co-routine has access to the shared stack only while running.
As such, local data must be stored in static variables (for co-routines that do not need to be re-entrant, \ie for which only a single instance will be running at any given time) or in memory allocated by the programmer for this purpose (\eg passed through co-routine's \texttt{args} parameter).
In practice, explicitly managing memory forces the programmer to reason about the lifetime of variables, ultimately encouraging clearer and more efficient programming.

\textbf{Comparison to GAP SDK.}
The difference between co-routines and the traditional multi-tasking model provided by the PMSIS GAP SDK is that \texttt{co\_event\_t} decouples the moment an event is initialized from the moment a co-routine waits for it.
Multiple waiters can all listen to the same event.
This allows a task to spawn sub-tasks, perform its work, and suspend itself when it finally needs the results, waiting on the sub-task completion events.
    \section{Experimental setup}
\label{app:experiments}

\begin{figure*}[t]
  \centering
  \includegraphics[width=\linewidth]{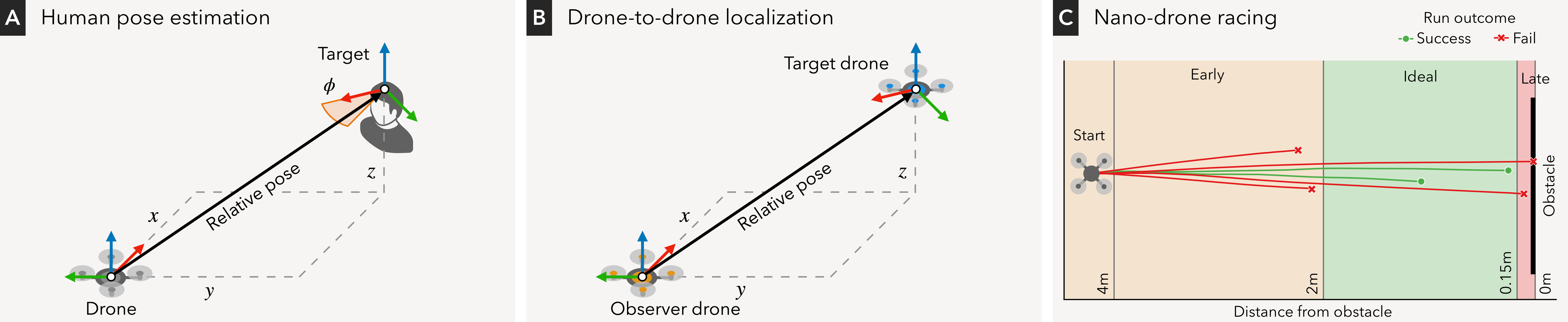}
  \caption{Autonomous tasks considered in our three in-field experiments: A) human pose estimation, B) drone-to-drone localization, and C) nano-drone racing.}
  \label{fig:tasks}
\end{figure*}

\begin{figure}[t]
  \centering
  \includegraphics[width=0.85\linewidth]{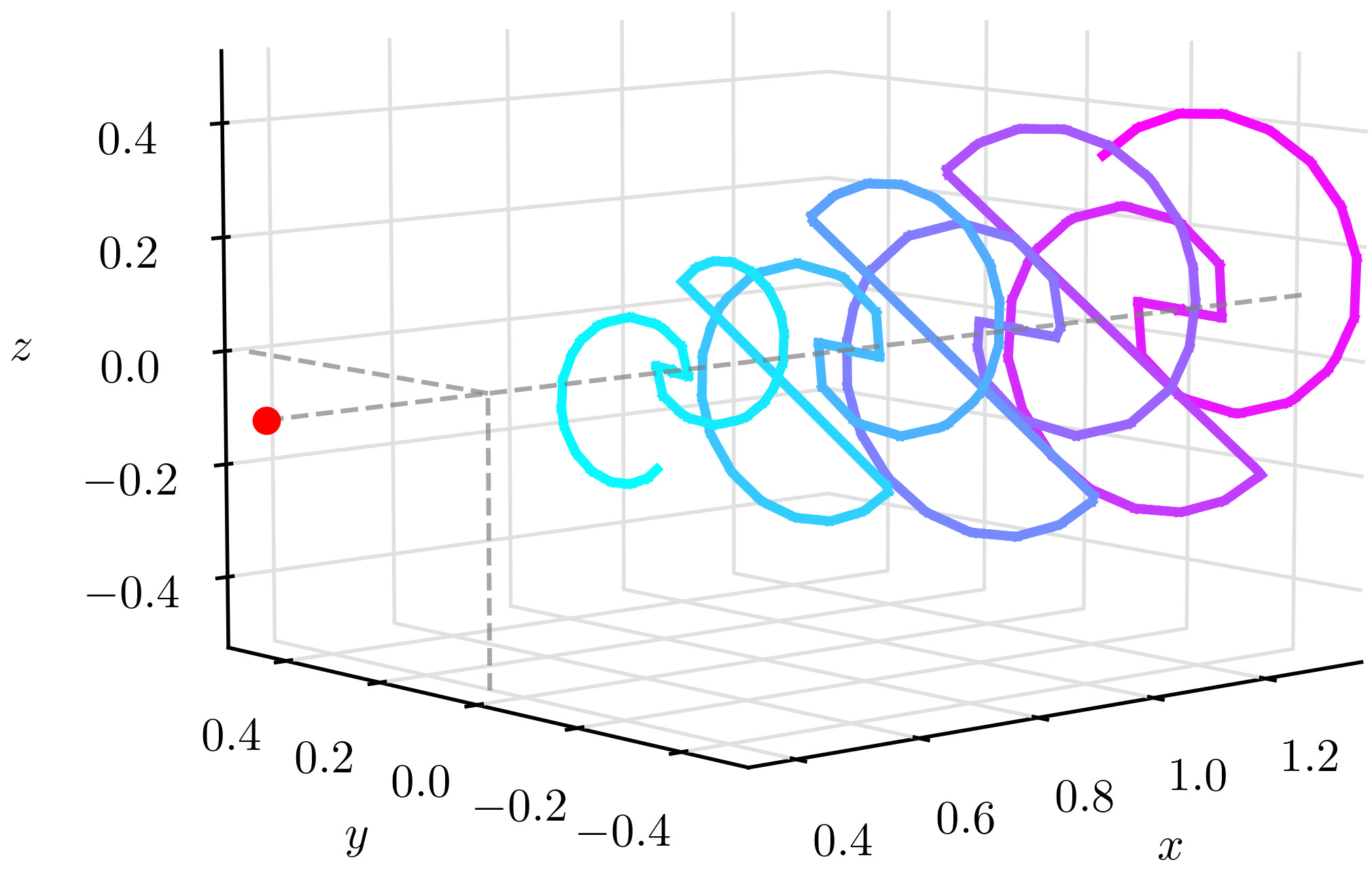}
  \caption{Three-dimensional trajectory flown by the target nano-drone during our drone-to-drone localization experiment~\cite{bonato2023d2d}.}
  \label{fig:d2d_spiral_trajectory}
\end{figure}

\subsection{Human pose estimation}
A convolutional neural network (CNN)~\cite{cereda2021improving,cereda2023secure} takes a single gray-scale $160\times\SI{96}{\pixel}$ camera frame as input, and outputs the subject's pose $(x, y, z, \phi)$ relative to drone's horizontal frame, depicted in Figure~\ref{fig:tasks}-A.
The estimated poses are filtered by a Kalman filter, and used to drive a ``follow-me'' velocity controller that keeps the drone in front of the subject, in the center of the image.

We evaluate our system with the experimental setup from the literature~\cite{pulp-frontnet}.
We track the subject's and drone's absolute poses along a pre-determined test trajectory in a mocap-equipped laboratory.
We report two metrics of control performance used in the literature. 
The \textit{mean horizontal position error} $e_\mathit{xy}$ represents the average distance in the XY plane of the drone from its desired position \SI{1.5}{\meter} in front of the subject.
The \textit{mean angular error} $e_{\theta}$ measures the average angle between the drone's orientation and its desired orientation (\ie looking directly at the subject).

\subsection{Drone-to-drone localization}
A fully convolutional neural network (FCNN)~\cite{crupi2024fcnn} takes a single gray-scale $160\times\SI{160}{\pixel}$ camera frame as input, and outputs three $20\times\SI{20}{\pixel}$ maps.
For each pixel, the first two maps represent the target's presence (position map) and distance from the observer (depth map), respectively.
The third map estimates the state of the target's LEDs, which are employed for different use cases and not used in this work.
We compute the target's image-plane 2D coordinates $(u, v)$ from the barycenter of the position map activations and the target's 3D distance $d$ from the average of the depth map weighed by the position map.
Finally, we convert the $(u, v)$ and $d$ coordinates into the target's pose $(x, y, z)$ relative to the observer's horizontal frame, depicted in Figure~\ref{fig:tasks}-B.

A closed-loop velocity controller then maintains the observer at a fixed distance $\Delta = \SI{0.8}{\meter}$ from the target drone, trying to keep it in the center of the image.
We follow the experimental setup of Crupi \textit{et al.}~\cite{crupi2024fcnn}, in which the target performs the pre-recorded spiral  trajectory shown in Figure~\ref{fig:d2d_spiral_trajectory}.
We evaluate the control performance through the \textit{mean position error} $e_\mathit{xyz}$, which measures the Euclidean distance between the observer's desired and actual position in 3D space w.r.t. the target drone.
This metric measures the observer's ability to track the target and reproduce its trajectory.

\subsection{Nano-drone racing}
We consider a drone autonomously exploring an environment with a CNN~\cite{lamberti2024imav} that predicts, for each input frame, the presence of obstacles within \SI{2}{\meter} of the drone.
The CNN takes as input a single gray-scale $162\times\SI{162}{\pixel}$ camera frame and produces three \textit{collision probability} scalar outputs, which represent respectively the presence of obstacles in the left, center, and right regions forward of the drone.
The collision probabilities are then filtered by a low-pass filter and used to bring the drone to a stop, when it comes in an obstacle's proximity.

We focus our evaluation on the system's obstacle detection performance.
The drone takes off facing an obstacle at a pre-determined \SI{4}{\meter} distance, and the controller drives it at a target forward speed toward the obstacle.
In our experiment, we track the drone's absolute pose with a mocap system, recording the minimum distance from the obstacle over the course of the run.
We evaluate the control performance in terms of \textit{success rate}, \ie the percentage of runs in which the drone successfully halts inside the ideal zone depicted in Figure~\ref{fig:tasks}-C.

While the drone is flying towards the obstacle, the controller breaks the drone as soon as the CNN's predicted collision probability rises above a threshold.
We experimentally select a separate breaking threshold (0.05 and 0.30) for each of the two throughput configurations (5 and \SI{30}{\hertz}, respectively) to achieve the maximum success rate.
Due to the reduced number of collision probability samples at \SI{5}{\hertz} a lower threshold is necessary compared to \SI{30}{\hertz} to ensure the drone breaks in time (the ideal zone in Figure~\ref{fig:tasks}-C).
Too low thresholds, on the other hand, cause the drone to break prematurely (the early zone).
To identify the best thresholds, we fly the drone in open loop toward the obstacle and record the CNN outputs without the predictions impacting the control.
We perform 18 experiment runs at max speeds 1.0 through \SI{3.5}{\meter/\second} and analyze the data offline to determine each run's outcome at each possible breaking threshold.

\begin{figure}[t]
  \centering
  \includegraphics[width=\columnwidth]{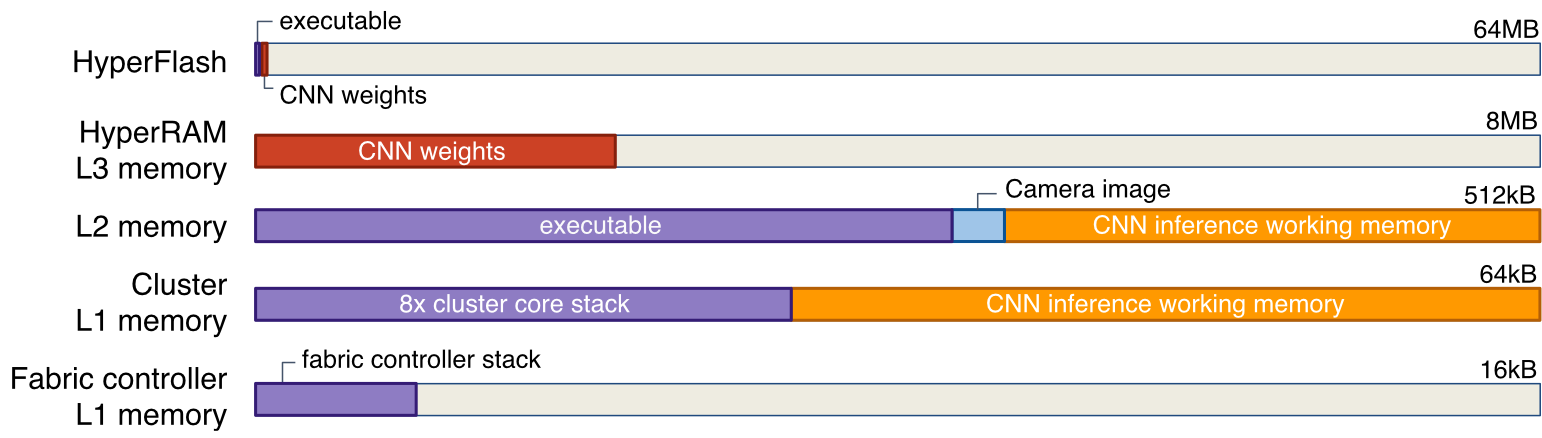}
  \caption{Memory breakdown of the deployed PULP-Frontnet CNN for the human pose estimation task.}
  \label{fig:memory}
\end{figure}

\subsection{Neural network deployment}
Across all three experiments, onboard CNNs are deployed on the GAP8 SoC through an automated pipeline.
We train a full-precision model in PyTorch with \texttt{float32} arithmetic.
We then quantize it to \texttt{int8} arithmetic with the open-source QuantLib~\cite{quantlib} quantization library through a quantization-aware fine-tuning post-training process.
This results in a $4\times$ reduction in memory footprint, with a marginal loss in performance metrics (\ie mean horizontal position and angular error for the human pose estimation task, mean position error for drone-to-drone localization, and success rate for nano-drone racing).

The DORY~\cite{dory} framework then automatically generates the C code for onboard model inference, \ie given a single input image in L2 memory, it computes and writes the model outputs back to L2 memory.
DORY relies on computation kernels from the PULP-NN-mixed~\cite{pulp-nn-mixed} library for individual model layers (\eg convolution, max pooling, etc.).
These hand-written routines are tuned to exploit GAP8's parallel processing, low-latency L1 memory, and single-instruction multiple-data (SIMD) vector instructions.
DORY's generated code orchestrates the computation kernels' execution, automatically subdivides each layer in subunits that fit the L1 memory (memory tiling), and schedules DMA memory transfers in parallel to computation.

Our NanoCockpit framework, on the other hand, implements all support software tasks required to run an autonomous application on a nano-drone, including camera acquisition and wireless streaming tasks.
For each application, a minimal user-provided \texttt{main.c} file integrates the automatically generated inference code on top of the NanoCockpit framework, resulting in a full closed-loop application.
Figure~\ref{fig:memory} analyzes the whole-system memory usage of the PULP-Frontnet CNN from the human pose estimation task in Section~\ref{sec:results}-A.

    \section{Discussion}
\label{app:discussion}

Our contribution represents a reference implementation of a performance-optimized application framework for ultra-constrained miniaturized robots, in addition to a solution closely coupled to the Crazyflie and AI-deck hardware.
The techniques and design patterns we present can be grouped in three categories, each with a different degree of transferability.
First, our GAP8-specific optimizations are tailored to this particular SoC and would require adaptation for newer processors. However, the underlying design principles, \eg double-buffered DMA transfers, cluster-level parallelism, and memory-aware task scheduling, remain applicable to similar architectures.
Second, our ESP32 communication stack targets one of the most widely adopted Wi-Fi modules in the embedded ecosystem, making it directly reusable on any platform that employs this chip.
Third, our heterogeneous CPS integration, which orchestrates collaboration between the STM32 flight controller, the ESP32 communication module, and the GAP8 mission controller, addresses architectural challenges common to a broad class of resource-constrained multi-MCU systems.

Concrete alternative hardware already exists within the Crazyflie ecosystem.
The GAP9Shield~\cite{muller2024gap9shield} is an open-source academic prototype built around the upgraded GAP9 Parallel Ultra-Low-Power (PULP) System-on-Chip.
Given the substantial hardware and software-stack similarities between the GAP8 and GAP9, our NanoCockpit framework can be ported to this newer platform with limited effort.
Additional camera deck prototypes, tailored for remote computation, have been demonstrated by Bitcraze\footnote{\url{https://www.bitcraze.io/2024/04/prototype-of-forward-facing-expansion-connector/}}.
These decks continue to rely on an ESP32 Wi-Fi module, identical to the one on the AI-deck, for high-bandwidth communication with the remote base station; consequently, our optimized communication stack can be leveraged directly from these custom decks as well.

More broadly, by releasing our framework as open source, we provide the research community with a reusable foundation that can be ported and adapted to future nano-robotics hardware and related embedded systems.

    \printbibliography

@article{pulp-frontnet,
  author={Palossi, Daniele and Zimmerman, Nicky and Burrello, Alessio and Conti, Francesco and Müller, Hanna and Gambardella, Luca Maria and Benini, Luca and Giusti, Alessandro and Guzzi, Jérôme},
  journal={IEEE Internet of Things Journal}, 
  title={Fully Onboard {AI}-powered Human-Drone Pose Estimation on Ultra-low Power Autonomous Flying Nano-{UAVs}}, 
  year={2021},
  volume={},
  number={},
  pages={1-1},
  doi={10.1109/JIOT.2021.3091643}
 }

@inproceedings{cereda2021improving,
  title={Improving the generalization capability of {DNNs} for ultra-low power autonomous nano-{UAVs}},
  author={Cereda, Elia and Ferri, Marco and Mantegazza, Dario and Zimmerman, Nicky and Gambardella, Luca M and Guzzi, J{\'e}r{\^o}me and Giusti, Alessandro and Palossi, Daniele},
  booktitle={2021 17th International Conference on Distributed Computing in Sensor Systems (DCOSS)},
  pages={327--334},
  year={2021},
  organization={IEEE}
}

@article{lamberti2024imav,
  author={Lamberti, Lorenzo and Cereda, Elia and Abbate, Gabriele and Bellone, Lorenzo and Morinigo, Victor Javier Kartsch and Barciś, Michał and Barciś, Agata and Giusti, Alessandro and Conti, Francesco and Palossi, Daniele},
  journal={IEEE Robotics and Automation Letters}, 
  title={A Sim-to-Real Deep Learning-Based Framework for Autonomous Nano-Drone Racing}, 
  year={2024},
  volume={9},
  number={2},
  pages={1899-1906},
  doi={10.1109/LRA.2024.3349814}
}

@INPROCEEDINGS{crupi2024fcnn,
  author={Crupi, Luca and Giusti, Alessandro and Palossi, Daniele},
  booktitle={2024 IEEE International Conference on Robotics and Automation (ICRA)}, 
  title={High-throughput Visual Nano-drone to Nano-drone Relative Localization using Onboard Fully Convolutional Networks}, 
  year={2024},
  volume={},
  number={},
  pages={5345-5351}
}

@inproceedings{cereda2023secure,
  author = {Cereda, Elia and Giusti, Alessandro and Palossi, Daniele},
  title = {Secure Deep Learning-based Distributed Intelligence on Pocket-sized Drones},
  year = {2023}, 
  publisher = {Association for Computing Machinery},
  address = {New York, NY, USA},
  booktitle = {Proceedings of the 2023 International Conference on Embedded Wireless Systems and Networks},
  pages = {409–414},
  numpages = {6},
  location = {Rende, Italy},
  series = {EWSN '23}
}

@article{niculescu2021improving,
  title={Improving autonomous nano-drones performance via automated end-to-end optimization and deployment of {DNNs}},
  author={Niculescu, Vlad and Lamberti, Lorenzo and Conti, Francesco and Benini, Luca and Palossi, Daniele},
  journal={IEEE Journal on Emerging and Select Topics in Circuits and Systems},
  volume={11},
  number={4},
  pages={548--562},
  year={2021},
  publisher={IEEE}
}

@article{chen23pedestrian,
  title={Fully Onboard Single Pedestrian Tracking On Nano-{UAV} Platform},
  author={Chen, Haolin and Wu, Ruidong and Lu, Wenshuai and Ji, Xinglong and Wang, Tao and Ding, Haolun and Dai, Yuxiang and Liu, Bing},
  journal={Journal of Intelligent \& Robotic Systems},
  volume={109},
  number={3},
  pages={50},
  year={2023},
  publisher={Springer}
}

@inproceedings{zhou23pedestrian,
  title={Solving Tracking Challenges: Lightweight Algorithm for Single Pedestrian Tracking on Nano-{UAVs}},
  author={Zhou, Jingtao and Feng, Lei and Chen, Haolin and Liu, Bing},
  booktitle={2023 IEEE 16th International Conference on Electronic Measurement \& Instruments (ICEMI)},
  pages={486--492},
  year={2023},
  organization={IEEE}
}

@INPROCEEDINGS{pourjabar23multi,
  author={Pourjabar, Mahyar and Rusci, Manuele and Bompani, Luca and Lamberti, Lorenzo and Niculescu, Vlad and Palossi, Daniele and Benini, Luca},
  booktitle={2023 30th IEEE International Conference on Electronics, Circuits and Systems (ICECS)}, 
  title={Multi-sensory Anti-collision Design for Autonomous Nano-swarm Exploration}, 
  year={2023},
  volume={},
  number={},
  pages={1-5},
  doi={10.1109/ICECS58634.2023.10382769}}

@inproceedings{nanoflownet,
  title={{NanoFlowNet}: Real-time dense optical flow on a nano quadcopter},
  author={Bouwmeester, Rik J and Paredes-Vall{\'e}s, Federico and De Croon, Guido CHE},
  booktitle={2023 IEEE International Conference on Robotics and Automation (ICRA)},
  pages={1996--2003},
  year={2023},
  organization={IEEE}
}

@inproceedings{navardi2022optimization,
  title={An optimization framework for efficient vision-based autonomous drone navigation},
  author={Navardi, Mozhgan and Shiri, Aidin and Humes, Edward and Waytowich, Nicholas R and Mohsenin, Tinoosh},
  booktitle={2022 IEEE International Conference on Artificial Intelligence Circuits and Systems (AICAS)},
  pages={304--307},
  year={2022},
  organization={IEEE}
}

@InProceedings{muller2024gap9shield,
author="M{\"u}ller, Hanna
and Kartsch, Victor
and Benini, Luca",
editor="Secchi, Cristian
and Marconi, Lorenzo",
title="{GAP9Shield}: A {150GOPS} {AI}-Capable Ultra-low Power Module for Vision and Ranging Applications on Nano-drones",
booktitle="European Robotics Forum 2024",
year="2024",
publisher="Springer Nature Switzerland",
address="Cham",
pages="292--297",
}

@INPROCEEDINGS{bompani23bio,
  author={Lamberti, Lorenzo and Bompani, Luca and Kartsch, Victor Javier and Rusci, Manuele and Palossi, Daniele and Benini, Luca},
  booktitle={2023 Design, Automation \& Test in Europe Conference \& Exhibition (DATE)}, 
  title={Bio-inspired Autonomous Exploration Policies with {CNN}-based Object Detection on Nano-drones}, 
  year={2023},
  volume={},
  number={},
  pages={1-6},
  doi={10.23919/DATE56975.2023.10137154}}

@INPROCEEDINGS{bonato2023d2d,
  author={Bonato, S. and Lambertenghi, S. C. and Cereda, E. and Giusti, A. and Palossi, D.},
  booktitle={2023 IEEE International Conference on Robotics and Automation (ICRA)}, 
  title={Ultra-low Power Deep Learning-based Monocular Relative Localization Onboard Nano-quadrotors}, 
  year={2023},
  volume={},
  number={},
  pages={3411-3417},
  doi={10.1109/ICRA48891.2023.10161127}}

@INPROCEEDINGS{zauli2024vibration,
  author={Zauli, Matteo and Pirazzi, Marco and Zonzini, Federica and De Marchi, Luca},
  booktitle={2024 IEEE Sensors Applications Symposium (SAS)}, 
  title={Exploiting Nano Aerial Vehicles as Sensor Nodes for Wireless Vibration Monitoring}, 
  year={2024},
  volume={},
  number={},
  pages={1-6},
  doi={10.1109/SAS60918.2024.10636630}
}

@ARTICLE{toumieh2024motion,
  author={Toumieh, Charbel and Floreano, Dario},
  journal={IEEE Transactions on Robotics}, 
  title={High-Speed Motion Planning for Aerial Swarms in Unknown and Cluttered Environments}, 
  year={2024},
  volume={40},
  number={},
  pages={3642-3656}
}

@INPROCEEDINGS{kumar2024watchers,
  author={Kumar, I Navin and Bandyopadhyay, Sudarshan and Reddy G, Preetham and Godkhindi, Shrutkirthi S. and Prabhakar, T V},
  booktitle={2024 16th International Conference on COMmunication Systems \& NETworkS (COMSNETS)}, 
  title={{Watcher of the Warehouse}: Edge-Based Low Power Inventory Management Using Nano Drones}, 
  year={2024},
  volume={},
  number={},
  pages={315-317},
  doi={10.1109/COMSNETS59351.2024.10426854}}

@ARTICLE{kalenberg2024stargate,
  author={Kalenberg, Konstantin and Müller, Hanna and Polonelli, Tommaso and Schiaffino, Alberto and Niculescu, Vlad and Cioflan, Cristian and Magno, Michele and Benini, Luca},
  journal={IEEE Internet of Things Journal}, 
  title={{Stargate}: Multimodal Sensor Fusion for Autonomous Navigation on Miniaturized {UAVs}}, 
  year={2024},
  volume={11},
  number={12},
  pages={21372-21390},
  doi={10.1109/JIOT.2024.3363036}}

@INPROCEEDINGS{mengozzi2024drl,
  author={Mengozzi, Sebastiano and Zanatta, Luca and Barchi, Francesco and Bartolini, Andrea and Acquaviva, Andrea},
  booktitle={2024 IEEE International Conference on Omni-layer Intelligent Systems (COINS)}, 
  title={Towards Nano-Drones Agile Flight Using Deep Reinforcement Learning}, 
  year={2024},
  volume={},
  number={},
  pages={1-6},
  doi={10.1109/COINS61597.2024.10622558}}

@InProceedings{kazim2024nmpc,
author={Kazim, Muhammad and Sim, Hyunjae and Shin, Gihun and Hwang, Hwancheol and Kim, Kwang-Ki K.},
editor={Lee, Soon-Geul and An, Jinung and Chong, Nak Young and Strand, Marcus and Kim, Joo H.},
title={Aggressive Trajectory Tracking for Nano Quadrotors Using Embedded Nonlinear Model Predictive Control},
booktitle={Intelligent Autonomous Systems 18},
year={2024},
publisher={Springer Nature Switzerland},
address={Cham},
pages={317--332},
isbn={978-3-031-44851-5}
}

@ARTICLE{navardi2023metae2rl,
  author={Navardi, Mozhgan and Humes, Edward and Manjunath, Tejaswini and Mohsenin, Tinoosh},
  journal={IEEE Micro}, 
  title={{MetaE2RL}: Toward Meta-Reasoning for Energy-Efficient Multigoal Reinforcement Learning With Squeezed-Edge You Only Look Once}, 
  year={2023},
  volume={43},
  number={6},
  pages={29-39},
  doi={10.1109/MM.2023.3318200}}

@INPROCEEDINGS{mueller2024bridges,
  author={Müller, David and Herbers, Patrick and Dyrska, Raphael and Çelik, Firdes and König, Markus and Mönnigmann, Martin},
  booktitle={2024 18th International Conference on Control, Automation, Robotics and Vision (ICARCV)}, 
  title={{Inside Bridges}: Autonomous Crack Inspection with Nano {UAVs} in {GNSS}-Denied Environments}, 
  year={2024},
  volume={},
  number={},
  pages={910-915},
  doi={10.1109/ICARCV63323.2024.10821644}}

@INPROCEEDINGS{sexton2022tbb,
  author={Sexton, Connor and Callenes, Joseph},
  booktitle={2022 52nd Annual IEEE/IFIP International Conference on Dependable Systems and Networks Workshops (DSN-W)}, 
  title={Tiny Black Boxes: A nano-Drone Safety Architecture}, 
  year={2022},
  volume={},
  number={},
  pages={12-19},
  doi={10.1109/DSN-W54100.2022.00013}}

@article{olejnik2020flappingservo,
author = {Olejnik, Diana A. and Duisterhof, Bardienus P. and Kar\'{a}sek, Matej and Scheper, Kirk Y. W. and van Dijk, Tom and de Croon, Guido C. H. E.},
title = {A Tailless Flapping Wing {MAV} Performing Monocular Visual Servoing Tasks},
journal = {Unmanned Systems},
volume = {08},
number = {04},
pages = {287-294},
year = {2020},
doi = {10.1142/S2301385020500235},
}

@ARTICLE{shao2022millipede,
  author={Shao, Qi and Dong, Xuguang and Lin, Zhonghan and Tang, Chao and Sun, Hao and Liu, Xin-Jun and Zhao, Huichan},
  journal={IEEE Robotics and Automation Letters}, 
  title={Untethered Robotic Millipede Driven by Low-Pressure Microfluidic Actuators for Multi-Terrain Exploration}, 
  year={2022},
  volume={7},
  number={4},
  pages={12142-12149}
}

@INPROCEEDINGS{kabutz2023mclari,
  author={Kabutz, Heiko and Hedrick, Alexander and McDonnell, William P. and Jayaram, Kaushik},
  booktitle={2023 IEEE/RSJ International Conference on Intelligent Robots and Systems (IROS)}, 
  title={{mCLARI}: A Shape-Morphing Insect-Scale Robot Capable of Omnidirectional Terrain-Adaptive Locomotion in Laterally Confined Spaces}, 
  year={2023},
  volume={},
  number={},
  pages={8371-8376},
  doi={10.1109/IROS55552.2023.10341588}}

@ARTICLE{goldberg2018legged,
  author={Goldberg, Benjamin and Zufferey, Raphael and Doshi, Neel and Helbling, Elizabeth Farrell and Whittredge, Griffin and Kovac, Mirko and Wood, Robert J.},
  journal={IEEE Robotics and Automation Letters}, 
  title={Power and Control Autonomy for High-Speed Locomotion With an Insect-Scale Legged Robot}, 
  year={2018},
  volume={3},
  number={2},
  pages={987-993},
  doi={10.1109/LRA.2018.2793355}}

@article{saeed2025flapping,
author = {Saeed Rafee Nekoo and Ramy Rashad and Christophe De Wagter and Sawyer B Fuller and Guido de Croon and Stefano Stramigioli and Anibal Ollero},
title ={A review on flapping-wing robots: Recent progress and challenges},
journal = {The International Journal of Robotics Research},
volume = {44},
number = {14},
pages = {2305-2339},
year = {2025},
doi = {10.1177/02783649251343638},
}

@INPROCEEDINGS {sartori2025autonomous,
    author = { Sartori, Mattia and Singhal, Chetna and Roy, Neelabhro and Brunelli, Davide and Gross, James },
    booktitle = { 2025 21st International Conference on Distributed Computing in Smart Systems and the Internet of Things (DCOSS-IoT) },
    title = {{ {AI} and Vision Based Autonomous Navigation of Nano-Drones in Partially-Known Environments }},
    year = {2025},
    volume = {},
    ISSN = {},
    pages = {307-314},
    doi = {10.1109/DCOSS-IoT65416.2025.00058},
    publisher = {IEEE Computer Society},
    address = {Los Alamitos, CA, USA},
    month =Jun
}

@article{belson2019coroutines,
author = {Belson, Bruce and Holdsworth, Jason and Xiang, Wei and Philippa, Bronson},
title = {A Survey of Asynchronous Programming Using Coroutines in the Internet of Things and Embedded Systems},
year = {2019},
issue_date = {May 2019},
publisher = {Association for Computing Machinery},
address = {New York, NY, USA},
volume = {18},
number = {3},
issn = {1539-9087},
doi = {10.1145/3319618},
journal = {ACM Trans. Embed. Comput. Syst.},
month = jun,
articleno = {21},
numpages = {21}
}

@book{embeddedRobotics,
  title={Embedded robotics},
  author={Br{\"a}unl, Thomas},
  year={2003},
  publisher={Springer}
}

@inproceedings{aideck,
  title={An open source and open hardware deep learning-powered visual navigation engine for autonomous nano-{UAVs}},
  author={Palossi, Daniele and Conti, Francesco and Benini, Luca},
  booktitle={2019 15th International Conference on Distributed Computing in Sensor Systems (DCOSS)},
  pages={604--611},
  year={2019},
  organization={IEEE}
}

@online{bitcraze2025cpx,
  author = {Bitcraze},
  title = {{CPX} - {Crazyflie Packet eXchange}},
  year = 2025,
  url = {https://web.archive.org/web/20250217104131/https://www.bitcraze.io/documentation/repository/crazyflie-firmware/master/functional-areas/cpx/},
  urldate = {2025-02-17}
}

@article{macenski2022ros2,
    author = {Steven Macenski and Tully Foote and Brian Gerkey and Chris Lalancette and William Woodall},
    title = {{Robot} {Operating} {System} 2: Design, architecture, and uses in the wild},
    journal = {Science Robotics},
    volume = {7},
    number = {66},
    pages = {eabm6074},
    year = {2022},
    doi = {10.1126/scirobotics.abm6074},
}

@manual{barry2024freertos,
  title        = {{FreeRTOS} Kernel},
  author       = {Richard Barry and the FreeRTOS Team},
  organization = {Real Time Engineers Ltd.},
  year         = {2025},
  url          = {https://www.freertos.org},
}

@manual{quantlib,
  title        = {QuantLib},
  author       = {Matteo Spallanzani and Georg Rutishauser and Moritz Scherer and Philip Wiese and Francesco Conti},
  year         = {2025},
  url          = {https://github.com/pulp-platform/quantlib},
  urldate      = {2025-02-17}
}

@article{dory,
    author={Burrello, Alessio and Garofalo, Angelo and Bruschi, Nazareno and Tagliavini, Giuseppe and Rossi, Davide and Conti, Francesco},
  journal={IEEE Transactions on Computers}, 
  title={{DORY}: Automatic End-to-End Deployment of Real-World {DNNs} on Low-Cost {IoT} {MCUs}}, 
  year={2021},
  volume={70},
  number={8},
  pages={1253-1268},
  doi={10.1109/TC.2021.3066883}
}

@inproceedings{pulp-nn-mixed,
author = {Bruschi, Nazareno and Garofalo, Angelo and Conti, Francesco and Tagliavini, Giuseppe and Rossi, Davide},
title = {Enabling Mixed-Precision Quantized Neural Networks in Extreme-Edge Devices},
year = {2020},
isbn = {9781450379564},
publisher = {Association for Computing Machinery},
address = {New York, NY, USA},
url = {https://doi.org/10.1145/3387902.3394038},
doi = {10.1145/3387902.3394038},
booktitle = {Proceedings of the 17th ACM International Conference on Computing Frontiers},
pages = {217–220},
numpages = {4},
location = {Catania, Sicily, Italy},
series = {CF ’20}
}
    \end{refsection}
\else
    \appendices

    \makeatletter
    \refstepcounter{section}  % Increments section number
    \@IEEEappendixsavesection*{}
    \addcontentsline{toc}{section}{}
    \label{app:related-work}

    \refstepcounter{section}  % Increments section number
    \@IEEEappendixsavesection*{}
    \addcontentsline{toc}{section}{}
    \label{app:coroutine}

    \refstepcounter{section}  % Increments section number
    \@IEEEappendixsavesection*{}
    \addcontentsline{toc}{section}{}
    \label{app:experiments}

    \refstepcounter{section}  % Increments section number
    \@IEEEappendixsavesection*{}
    \addcontentsline{toc}{section}{}
    \label{app:discussion}
    \makeatother
\fi

\end{document}